\documentclass[letterpaper, 10 pt, conference]{ieeeconf}  

\IEEEoverridecommandlockouts                              

\usepackage{graphics} 
\usepackage{epsfig} 
\usepackage{mathptmx} 
\usepackage{times} 
\usepackage{amsmath} 
\usepackage{amssymb}  

\usepackage{graphicx}
\usepackage{epstopdf}

\title{\LARGE \bf
CCDepth: A Lightweight Self-supervised Depth Estimation Network with Enhanced Interpretability
}

\author{Xi ZHANG$^{1}$, Yaru XUE$^{1, *}$, Shaocheng JIA$^{2, *}$, and Xin PEI$^{3}$
\thanks{This work has been accepted by The 27th IEEE International Conference on Intelligent Transportation Systems (IEEE ITSC 2024).}
\thanks{$^{1}$Xi ZHANG and Yaru XUE are with the College of Information Science and Engineering, China University of Petroleum (Beijing), Beijing, China.
}%
\thanks{$^{2}$Shaocheng JIA is with the Department of Civil Engineering, The University of Hong Kong, Hong Kong, China.
}%
\thanks{$^{3}$Xin PEI is with the Department of Automation, Beijing National Research Center for Information Science and Technology, Tsinghua University, Beijing, China and Shanghai Artificial Intelligence Laboratory, Shanghai, China.
}%
\thanks{*Corresponding authors. Email: {\tt\small xueyaru@cup.edu.cn}, {\tt\small shaocjia@connect.hku.hk}.
}%
}

\begin{document}

\maketitle
\thispagestyle{empty}
\pagestyle{empty}

\begin{abstract}

Self-supervised depth estimation, which solely requires monocular image sequence as input, has become increasingly popular and promising in recent years. Current research primarily focuses on enhancing the prediction accuracy of the models. However, the excessive number of parameters impedes the universal deployment of the model on edge devices. Moreover, the emerging neural networks, being black-box models, are difficult to analyze, leading to challenges in understanding the rationales for performance improvements. To mitigate these issues, this study proposes a novel hybrid self-supervised depth estimation network, CCDepth, comprising convolutional neural networks (CNNs) and the white-box CRATE (Coding RAte reduction TransformEr) network. This novel network uses CNNs and the CRATE modules to extract local and global information in images, respectively, thereby boosting learning efficiency and reducing model size. Furthermore, incorporating the CRATE modules into the network enables a mathematically interpretable process in capturing global features. Extensive experiments on the KITTI dataset indicate that the proposed CCDepth network can achieve performance comparable with those state-of-the-art methods, while the model size has been significantly reduced. In addition, a series of quantitative and qualitative analyses on the inner features in the CCDepth network further confirm the effectiveness of the proposed method.

\end{abstract}

\section{Introduction}

Depth estimation aims to predict depth map from one or multiple images, serving as an essential way for computers to understand and perceive the real world \cite{c1}. Accurately estimating depth maps plays a significant role in various applications, such as autonomous driving \cite{c2}, \cite{c3}, three-dimensional (3D) reconstruction, and robot vision \cite{c4}, \cite{c5}, \cite{c6}, \cite{c7}.

Current research on depth estimation predominantly focuses on designing models to improve accuracy, regardless of the continuous increase in model size, e.g., certain depth estimation models being tens or even hundreds of megabytes. However, such large models face challenges in practical deployment on edge devices, e.g., vehicles. To mitigate this issue, lightweight depth estimation models with comparable prediction performance are necessary to be developed.

Moreover, popular black-box models, including CNNs and vision Transformer (ViT), have been increasingly questioned in recent years. Specifically, the black-box models are faced with challenges in uninterpretable overfitting, arbitrary label generation, and most importantly, the vulnerability to adversarial attacks. For instance, Engstrom et al. demonstrated the drastic performance degradation of current black-box models even with minor perturbations \cite{c8}. Similarly, Szegedy et al. revealed that imposing subtle perturbations can lead neural networks to make erroneous predictions \cite{c9}. These limitations make the public not fully trust autonomous driving even artificial intelligence.

This study aims to mitigate the above-mentioned two issues by proposing a novel hybrid network, CCDepth, which comprises a white-box transformer CRATE and CNNs. Adopting an encoder-decoder structure, CNNs are used to capture fine local features in high-resolution images while the CRATE layers are used to extract global information. With the subtly designed CCDepth network, model size has been significantly reduced. Furthermore, the model’s interpretability is enhanced leveraging the interpretable CRATE components.

The rest of this paper is organized as follows. Section II reviews the related works. Section III details the proposed model. Sections IV and V report experimental results and conclude the paper, respectively.

\section{Related work}

\subsection{Depth estimation}

Depth estimation has attracted much attention due to its wide applications. Initially, hierarchical and multiscale Markov random field (MRF) was used for depth estimation \cite{c10}. Following that, many methods were developed for extracting image features based on expert knowledge, such as depth analogy \cite{c11}, light flow analysis \cite{c12}, etc.

With the rapid development of deep learning \cite{c13}, \cite{c14}, \cite{c15}, \cite{c16}, Eigen et al. applied deep learning to depth estimation tasks \cite{c17}. Since the advent of ResNet \cite{c18}, CNNs have been largely used in computer vision tasks \cite{c19}, \cite{c20}. In depth estimation, Monodepth \cite{c21} and Monodepth2 \cite{c22} were proposed and achieved excellent performance. Thereafter, many depth estimation networks have been proposed to conduct depth estimation \cite{c23}, \cite{c24}, \cite{c25}, \cite{c26}.

\begin{figure*}[thpb]
\centering
\includegraphics[scale=0.49]{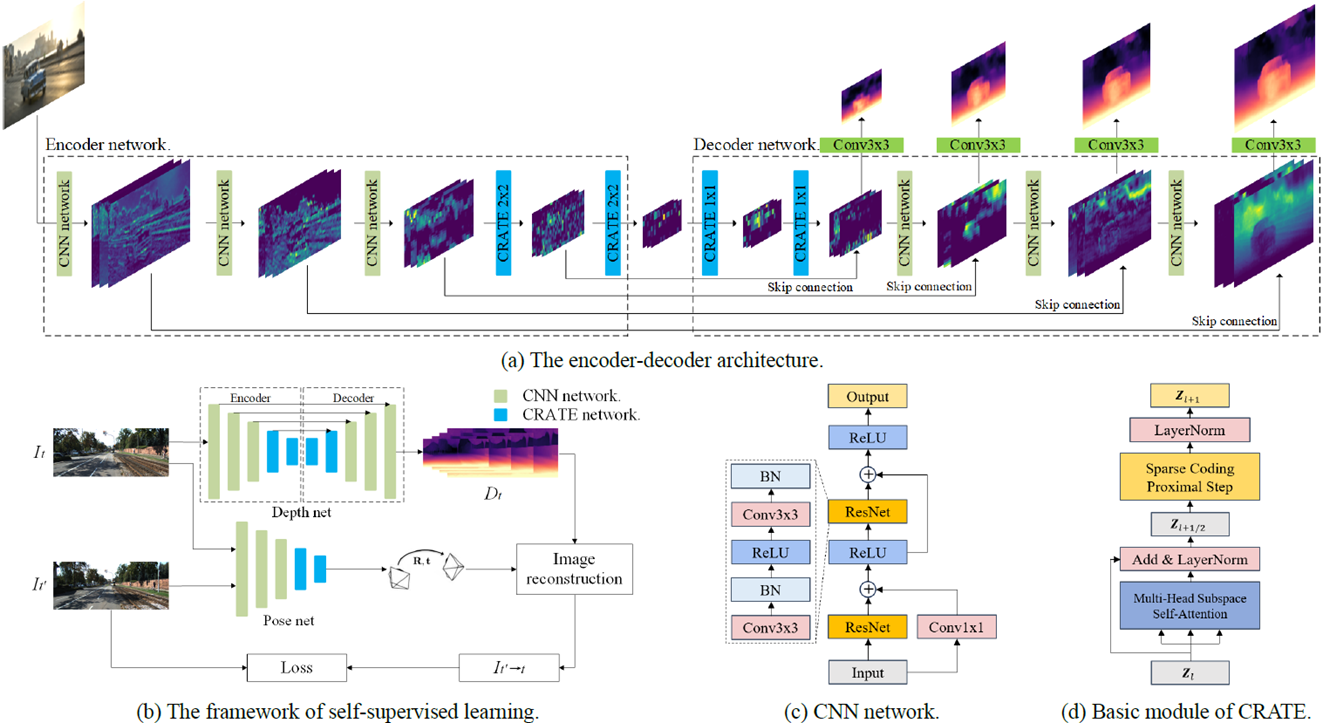}
\caption{The overall architecture of CCDepth.}
\label{figurelabel1}
\end{figure*}

\begin{figure}[thpb]
\centering
\includegraphics[scale=0.42]{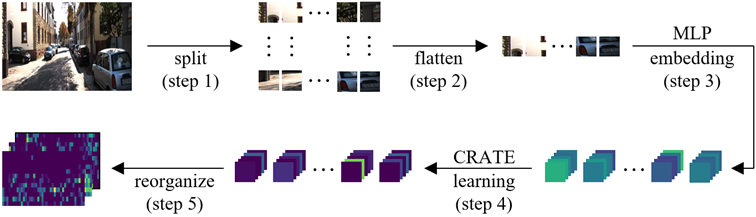}
\caption{Workflow of the CRATE layer. Step 1: the input image is split into several patches. Step 2: the two-dimensional (2D) image is flattened to a sequence. Step 3: each patch is embedded as a vector by Multilayer perceptron (MLP). Step 4: CRATE takes those embedded feature as input to learn more abstract features. Step 5: the 2D structure of the image is recovered.}
\label{figurelabel2}
\end{figure}

\subsection{Lightweight depth estimation model}
To reduce the number of parameters in depth estimation models, diverse methods have been adopted, such as using pyramid features \cite{c27}, recurrent neural networks (RNN) \cite{c28}, and a non-local coplanarity constraint and a novel attention mechanism (DAV) \cite{c29}. Moreover, Jia et al. extracted global information of images using the Fourier transform and proposed a joint learning of frequency and spatial domains \cite{c30}.

In summary, previous works have proposed a variety of depth estimation models and achieved substantial improvements in depth estimation accuracy. However, the depth estimation models still have a large number of parameters. The model’s interpretability is rarely explored, either. To mitigate the said issues, this work proposes a novel, hybrid, and lightweight network, CCDepth, for accurately predicting depth maps with enhanced interpretability.

\section{Method}
\subsection{Model architecture}
In this section, the architecture of the proposed depth estimation network, as shown in Figure \ref{figurelabel1}, is presented. The network features a U-Net structure. The resolution of the image/features is half reduced passing through each layer of the encoder, and double increased for the decoder. CNNs are used to extract the local information in high-resolution images while CRATE is used to extract global information in low-resolution images. 

Moreover, the model is trained with a self-supervised learning approach as shown in Figure \ref{figurelabel1}(b), where the depth net is used to predict depth information from the RGB image and the pose net estimates the movement between two consecutive frames. With the depth map and movement information, the target image can be reconstructed from the given reference images. Comparing the constructed target image with the actual target image, $I_{t'}$, offers the training loss.

\subsection{Encoder of CCDepth} 
For the sake of narration, we named the 10 layers of CCDepth as layer 1 to layer 10 respectively. The RGB image input received by the network is denoted as $\textbf{X}_{0}$, the output of layer ${i}$ is $\textbf{X}_{i}$.

The CNNs consist of two residual blocks, as shown in Figure \ref{figurelabel1}(c). The mirror reflection padding is adopted in CNNs. Given the input data of CNNs, $\textbf{X}_{i} \in \mathbb{R}^{h_{i} \times w_{i} \times C_{i}}$, $i = 0, 1, 2$, the output $\textbf{X}_{i+1} \in \mathbb{R}^{\frac{h_{i}}{2} \times \frac{w_{i}}{2} \times C_{i+1}}$, is given by
$$
\textbf{X}_{i+1/2}=\mathrm{CNN}(\textbf{X}_{i}), \eqno{(1)}
$$
$$
\textbf{X}_{i+1}=\mathrm{Maxpooling}(\textbf{X}_{i+1/2}), \eqno{(2)}
$$
where $\textbf{X}_{i+1/2} \in \mathbb{R}^{{h_{i}} \times w_{i} \times {C_{i+1}}}$ is the intermediate feature in the layer, $C_{i}$ and $C_{i+1}$ are the number of input and output image channels, respectively. After passing through three-layer CNNs, the CRATE layers are then introduced.

As shown in Figure \ref{figurelabel2}, the work of the CRATE layer is divided into five steps: in step 1, the input image is split into a number of patches, and the size of the split is determined by the parameter ‘patch size’; in step 2, the individual patches are flattened into a sequence; in step 3, each patch is transformed into a vector by patch embedding and fed into the CRATE model; in step 4, vectors are updated as they learn; in step 5, the vectors are reorganized into the format of step 1 to get the output image. The number of vectors is the number of pixels in the output image, and the length of the vector is the number of channels.

CRATE aims to find a mapping that transforms input features with nonlinear and multimodal distributions to (piecewise) linear and compact features. Each layer receives an input $\textbf{X}_{i} \in \mathbb{R}^{h_{i} \times w_{i} \times C_{i}}$, $i=3, 4$, from the previous layer and transforms it into a number of patches to get the data $\textbf{Z}_{1} \in \mathbb{R}^{d \times N}$, $d$ and $N$ are the length and number of vectors, respectively. After $L$ basic modules of CRATE as shown in Figure \ref{figurelabel1}(d), the output of a CRATE layer, $\textbf{Z}$, is obtained.
$$
\textbf{X}_{i} \rightarrow \textbf{Z}_{1} \rightarrow \textbf{Z}_{2} \rightarrow \cdots \rightarrow \textbf{Z}_{L+1} = \textbf{Z}, \eqno{(3)}
$$
where $L$ is set to 2 in this paper, meaning that each CRATE block consists of two basic modules as shown in Figure \ref{figurelabel1}(d). Reconstructing $\textbf{Z}$ back to the 2D image form according to Figure \ref{figurelabel2}, the output $\textbf{X}_{i+1}$ is derived.

In order to maximize the information gain and ensure that the output has a more compact encoding, the following objective function is proposed \cite{c31}:
$$
\max_{f \in F} \mathbb{E}_{\mbox{\footnotesize$\textbf{Z}$}}\big{[}\Delta R(\textbf{Z}; \textbf{U}_{[K]})-\lambda_{c}\|\textbf{Z}\|_{0}\big{]} 
$$
$$
=\max_{f \in F} \mathbb{E}_{\mbox{\footnotesize$\textbf{Z}$}}\big{[}R(\textbf{Z})-\sum\nolimits_{k=1}^{K} R(\textbf{U}_{k}^{T} \textbf{Z}) - \lambda_{c}\|\textbf{Z}\|_{0}\big{]}, 
\eqno{(4)}
$$
where $\Delta R(\textbf{Z}; \textbf{U}_{[K]})$ represents the rate reduction, $R(\textbf{Z})$ is the coding rate and is defined as
$$
R(\textbf{Z}) \dot{=} \frac{1}{2} \mathrm{logdet}(\textbf{I} + \frac{d}{N\epsilon^{2}} \textbf{Z} \textbf{Z}^{T});
\eqno{(5)}
$$
$\textbf{U}_{[K]}=(\textbf{U}_{k})_{k=1}^{K}$, $\textbf{U}_{k} \in \mathbb{R}^{d \times p}$ is the orthonormal basis, $K$ is the number of bases in $\textbf{U}_{[K]}$. Each basic block in CRATE takes $\textbf{Z}_{l} \in \mathbb{R}^{d \times N}$ as input and processes $\textbf{Z}_{l}$ through a Multi-Head Subspace Self-Attention (MSSA) block and an Iterative Shrinkage-Thresholding Algorithms (ISTA) block, yielding output $\textbf{Z}_{l+1}$. Notably, the MSSA and ISTA blocks are used in CRATE rather than the Multi-Head Attention and Feed-Forward Network (FFN) blocks used in ViT.

\begin{table*}[ht]
\caption{Quantitative results of self-supervised depth estimation on the KITTI dataset. SC represents space complexity. The best performances are marked \textbf{bold}. The second-best performances are \underline{underlined}.}
\begin{center}
\renewcommand\arraystretch{1.1}
\setlength{\tabcolsep}{2.5mm}
\begin{tabular}{c c c c c c c c c}
    \hline
    Models & \multicolumn{4}{c}{Errors $\downarrow$} & \multicolumn{3}{c}{Errors $\uparrow$} & SC \\
    \cline{2-8}
    ~ & Abs Rel & Sq Rel & RMSE & RMSE log & $\delta < 1.25$ & $\delta < 1.25^2$ & $\delta < 1.25^3$ & ~ \\
    \hline
    Zhou et al. \cite{c32} & 0.208 & 1.768 & 6.856 & 0.283 & 0.678 & 0.885 & 0.957 & 126.0M \\
    Geonet \cite{c33} & 0.153 & 1.328 & 5.737 & 0.232 & 0.802 & 0.934 & 0.972 & 229.3M \\
    Casser et al. \cite{c34} & 0.141 & 1.138 & 5.521 & 0.219 & 0.820 & 0.942 & 0.976 & 67.0M \\
    Monodepth2 \cite{c22} & \textbf{0.115} & 0.903 & 4.863 & \underline{0.193} & \textbf{0.877} & \textbf{0.959} & \underline{0.981} & 59.4M \\
    FSLNet-L \cite{c30} & 0.128 & 0.897 & 4.905 & 0.200 & 0.852 & 0.953 & 0.980 & \underline{16.5M} \\
    CNN-ViT & \underline{0.119} & \underline{0.857} & \underline{4.789} & 0.194 & 0.867 & \underline{0.958} & \underline{0.981} & 17.4M \\
    CCDepth (Our) & \textbf{0.115} & \textbf{0.830} & \textbf{4.737} & \textbf{0.190} & \underline{0.874} & \textbf{0.959} & \textbf{0.982} & \textbf{12.6M} \\
    \hline
\end{tabular}
\end{center}
\label{table 1}
\end{table*}

\begin{figure*}[thpb]
\centering
\includegraphics[scale=0.57]{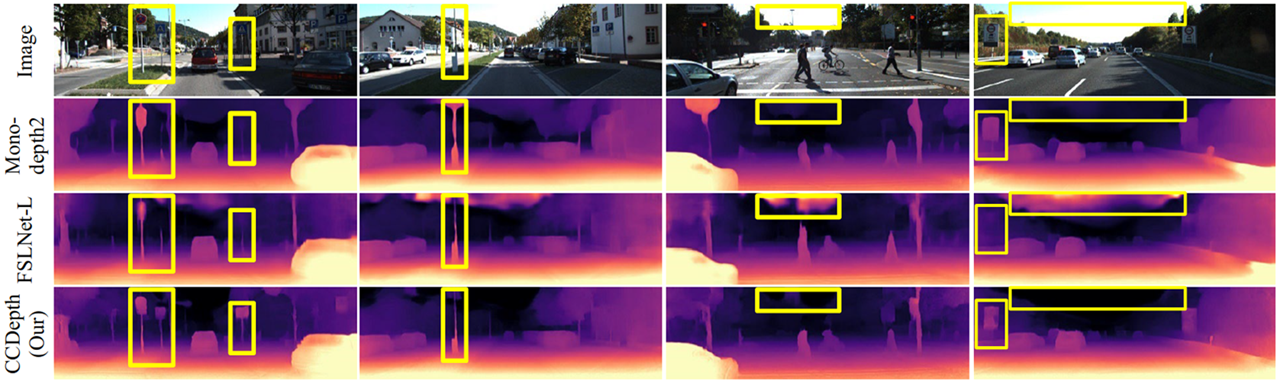}
\caption{Qualitative results of self-supervised depth estimation. (Monodepth2 \cite{c22}; FSLNet \cite{c30}).}
\label{figurelabel3}
\end{figure*}

\begin{table*}[ht]
\caption{Ablation studies on different scales. The best performances are marked \textbf{bold}.}
\begin{center}
\setlength{\tabcolsep}{2.5mm}
\begin{tabular}{c c c c c c c c}
    \hline
    Sacles & \multicolumn{4}{c}{Errors $\downarrow$} & \multicolumn{3}{c}{Errors $\uparrow$} \\
    \cline{2-8}
    ~ & Abs Rel & Sq Rel & RMSE & RMSE log & $\delta < 1.25$ & $\delta < 1.25^2$ & $\delta < 1.25^3$\\
    \hline
    1 & 0.119 & 0.860 & 4.750 & 0.194 & 0.870 & 0.958 & 0.981 \\
    2 & 0.121 & 0.861 & 4.760 & 0.195 & 0.866 & 0.957 & 0.981 \\
    3 & 0.118 & 0.841 & 4.738 & 0.194 & 0.870 & 0.958 & 0.981 \\
    4 & \textbf{0.115} & \textbf{0.830} & \textbf{4.737} & \textbf{0.190} & \textbf{0.874} & \textbf{0.959} & \textbf{0.982}\\
    \hline
\end{tabular}
\end{center}
\label{table 2}
\end{table*}

\begin{table*}[ht]
\caption{Ablation studies on different padding modes. The best performances are marked \textbf{bold}.}
\begin{center}
\setlength{\tabcolsep}{2.5mm}
\begin{tabular}{c c c c c c c c}
    \hline
    Padding mode & \multicolumn{4}{c}{Errors $\downarrow$} & \multicolumn{3}{c}{Errors $\uparrow$} \\
    \cline{2-8}
    ~ & Abs Rel & Sq Rel & RMSE & RMSE log & $\delta < 1.25$ & $\delta < 1.25^2$ & $\delta < 1.25^3$\\
    \hline
    zeros & 0.118 & 0.858 & 4.755 & 0.193 & 0.870 & 0.958 & 0.981 \\
    reflect & \textbf{0.115} & \textbf{0.830} & \textbf{4.737} & \textbf{0.190} & \textbf{0.874} & \textbf{0.959} & \textbf{0.982}\\
    \hline
\end{tabular}
\end{center}
\label{table 3}
\end{table*}

MSSA and ISTA each optimize a part of (4) so that the entire CRATE eventually achieves the desired effect of (4). The MSSA block is to optimize $\min_{\mbox{\footnotesize$\textbf{Z}$}} \sum_{k=1}^{K}R(\textbf{U}_{k}^{T}\textbf{Z})$,
$$
\textbf{Z}_{l+1/2}=\textbf{Z}_{l} - \kappa \nabla_{\mbox{\footnotesize$\textbf{Z}$}} \sum\nolimits_{k=1}^{K}R(\textbf{U}_{k}^{T}\textbf{Z})
$$
$$
\approx(1 - \kappa \cdot \frac{p}{N\epsilon^2})\textbf{Z}_l + \kappa \cdot \frac{p}{N\epsilon^2} \cdot \mathrm{MSSA}(\textbf{Z}_l|\textbf{U}_{[K]}),
\eqno{(6)}
$$
where $\epsilon$ is the precision of encoding, $\textbf{Z}_{l+1/2}$ is the output of MSSA block, an intermediate feature of $\textbf{Z}_l$ and $\textbf{Z}_{l+1}$. Applying the module to the subspace self-attention module offers
$$
\mathrm{SSA}(\textbf{Z}|\textbf{U}_{[K]}) \dot{=} (\textbf{U}_{k}^T\textbf{Z})\mathrm{softmax}\big{(}(\textbf{U}_k^T\textbf{Z})^T(\textbf{U}_k^T\textbf{Z})\big{)}.
\eqno{(7)}
$$
This further provides the MSSA operator
$$\mbox{\small $
\mathrm{MSSA}(\textbf{Z}|\textbf{U}_{[K]}) 
\dot{=}\frac{p}{N\epsilon^2}
\left[\begin{array}{ccc}
\textbf{U}_1,
\cdots,
\textbf{U}_K
\end{array}\right] 
\left[\begin{array}{ccc}
\mathrm{SSA}(\textbf{Z}|\textbf{U}_1)\\
\vdots\\
\mathrm{SSA}(\textbf{Z}|\textbf{U}_K)   
\end{array}\right]$}.      
\eqno{(8)}
$$
By replacing the Multi-Head Attention block in ViT, the module becomes a white box that acts as a compression (denoising) function.

The ISTA block is to planning $\max_{\mbox{\footnotesize$\textbf{Z}$}} R(\textbf{Z}) - \lambda_c \|\textbf{Z}\|_{0}$, First, set a (complete) incoherent or orthogonal dictionary $\textbf{D}$, such that $\textbf{Z}_{l+1/2}=\textbf{D}\textbf{Z}_{l+1}$. According to the incoherence assumption, $\textbf{D}^T\textbf{D} \approx \textbf{I}$, the following can be obtained:
$$
\textbf{Z}_{l+1}=\mathrm{argmin}\|\textbf{Z}\|_{0}, s.t.\, \textbf{Z}_{l+1/2}=\textbf{DZ}.
\eqno{(9)}
$$
Then, it follows
$$
\textbf{Z}_{l+1}=\mathrm{ReLU}\big{(}\textbf{Z}_{l+\frac{1}{2}}+\eta_{c}\textbf{D}^T(\textbf{Z}_{l+\frac{1}{2}}-\textbf{D}\textbf{Z}_{l+\frac{1}{2}})-\eta_{c}\lambda_1\big{)} 
$$
$$
\dot{=}\mathrm{ISTA}(\textbf{Z}_{l+1/2}|\textbf{D}),
\eqno{(10)}
$$
where $\eta_c > 0$ and $\lambda_1 > 0$ represent step size and sparsification regularizer.

Replacing the FFN block in ViT with an ISTA block furnishes data sparsification and completes the CRATE. Additionally, each MSSA includes 6 SSA heads. In the encoder network, patch size is set to 2. When an input data $\textbf{X}_i \in \mathbb{R}^{h_i \times w_i \times C_i}$ is fed into the layer,
$$
\textbf{X}_{i+1}=\mathrm{CRATE}_{2 \times 2}(\textbf{X}_i) \in \mathbb{R}^{\frac{h_i}{2} \times \frac{w_i}{2} \times C_{i+1}},
\eqno{(11)}
$$
where $i=3, 4$.

\subsection{Decoder of CCDepth}
The structure of the decoder network is symmetric with the encoder part. The difference is that the patch size of the CRATE network is set to 1 to keep the original resolution. To recover the image resolution, an upsampling operation is adopted as follows.
$$
\textbf{X}_{i+1/2}=\mathrm{CRATE}_{1 \times 1}(\textbf{X}_i) \in \mathbb{R}^{h_i \times w_i \times C_{i+1}},
\eqno{(12)}
$$
$$
\textbf{X}_{i+1}=\mathrm{Upsampling}(\textbf{X}_{i+1/2}) \in \mathbb{R}^{2h_i \times 2w_i \times C_{i+1}},
\eqno{(13)}
$$
where $i=5, 6$. After a 5-stage decoding process, the features are recovered to the original image resolution. For prediction, a convolutional layer with a size of $3 \times 3$ kernel is used as prediction head, as shown below.
$$
\textbf{D}_t = \mathrm{Conv}_{3 \times 3}(\textbf{x}_i \in \mathbb{R}^{h_i \times w_i \times C_i}), i=7, ..., 10,
\eqno{(14)}
$$
where $\textbf{D}_t \in \mathbb{R}^{h_i \times w_i \times 1}$.

\subsection{Loss function}
The training loss of the CCDepth network is a weighted summation of the photometric loss and the smoothness loss. The photometric loss is defined as
$$
L_p = \sum\nolimits_{t'}{pe(I_{t}, I_{t' \rightarrow t})},
\eqno{(15)}
$$
$$
pe(I_{t}, I_{t' \rightarrow t}) = \frac{\alpha}{2}\big{(}1 - \mathrm{SSIM}(I_{t}, I_{t' \rightarrow t})\big{)} 
$$
$$+ (1 - \alpha)\|I_t - I_{t' \rightarrow t}\|_{1},
\eqno{(16)}
$$
where $\mathrm{SSIM}$ represents the structure similarly loss; $\alpha = 0.85$; $I_t$ is the original image in the dataset; $I_{t'} \in \{I_{t-1}, I_{t+1}\}$ is the reference frame of $I_t$; and $I_{t' \rightarrow t}$ is the reconstructed target image from the reference image. Simultaneously, an automatic mask, as stated below, is used to mask out stationary pixels between consecutive frames.
$$
\mu = \big{[}\min_{t'}{pe(I_{t}, I_{t' \rightarrow t})} < \min_{t'}{pe(I_{t}, I_{t'})}\big{]},
\eqno{(17)}
$$
where $pe(\cdot, \cdot)$ represents the function to calculate the photometric loss.

Moreover, an edge-aware smoothness loss is adopted to encourage a smooth depth map.
$$
L_s = |\partial_x d_t^*|e^{-|\partial_xI_t|} + |\partial_y d_t^*|e^{-|\partial_yI_t|},
\eqno{(18)}
$$
where $d_t^* = d_t / \overline{d_t}$ is the mean-normalized inverse depth; $\partial_x$ and $\partial_y$ represent the differentiation operations along the $x$ and $y$ directions, respectively. The total loss is given by
$$
L_n = \mu L_p + \lambda L_s,
\eqno{(19)}
$$
where $\lambda$ is set to $0.001$ in this paper. With multi-scale prediction, the final loss is computed as
$$
L = \sum\nolimits_{n=1}^{N} {\frac{L_n}{2^{n - 1}}},
\eqno{(20)}
$$
where $L_n$ is the loss from the $n^{th}$ scale.

\section{Experiments}
In this section, comprehensive experiments are conducted to demonstrate the effectiveness of the proposed model. All experiments are implemented with PyTorch 1.10.2 on a single 2080Ti GPU card.

\subsection{Dataset}
In this paper, the KITTI dataset \cite{c35} is used for model training and evaluation. The Eigen’s split is adopted \cite{c36}.The static frames are removed following the preprocessing method in \cite{c32}. This results in 39,810 and 4,424 images for training and validation, respectively. The test set contains 697 images that were used for quantitative and qualitative analysis. All models are trained for 20 epochs using Adam optimization algorithm \cite{c37}. The learning rate is initialized at $10^{-4}$ and subsequently reduced to $10 ^ {-5}$ after 15 epochs. The batch size is set to 8. The input image resolution is set to $640 \times 192$.

\subsection{Results}
Quantitative results of self-supervised depth estimation are reported in Table \ref{table 1}. The proposed CCDepth model achieved performance competitive on most metrics to those of state-of-the-art methods. In particular, the number of parameters in CCDepth is only 12.6M, which is 78.8$\%$ and 23.6$\%$ less than that in Monodepth2 and FSLNet. This clearly demonstrates the effectiveness and efficiency of the proposed model.

Some qualitative results are shown in Figure \ref{figurelabel3}. The test images are from the KITTI dataset. It further confirms that the proposed model outperformed other models, especially in thin objects and sky regions. In addition, the computation time of CCDepth for predicting a single image is only 15.71 ms on the designated machine, which can meet the requirement of most real-time applications.

The observed advancements may benefit from the hybrid network, which can effectively extract both local and global features. Specifically, only several convolutional layers are adequate to capture the crucial details in the image. Furthermore, the CRATE layers applied to the low-resolution features can efficiently extract global information, thereby significantly improving depth estimation performance.

\subsection{Ablation studies}
The ablation studies on different numbers of prediction scales and padding modes are further investigated. 

The results under different prediction scales are presented in Table \ref{table 2}. It is found that the performance tends to gradually decrease as the number of scales decreases. Therefore, four-scale prediction is selected.

\begin{figure}[thpb]
\centering
\includegraphics[scale=0.52]{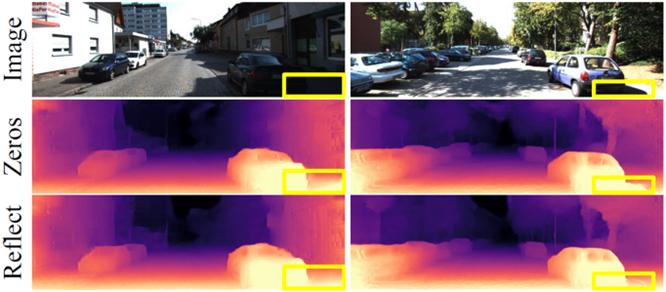}
\caption{Qualitative results of self-supervised depth estimation with different padding modes.}
\label{figurelabel4}
\end{figure}

The results under different padding modes are presented in Table \ref{table 3}. The reflect padding performed much better than zero padding. Figure \ref{figurelabel4} presents some qualitative results. It can be found that models trained with zero-padding exhibited noticeable distortions in some edge areas. The reason for this phenomenon can be explained by the fact that when zero-padding is applied, the zero elements used for padding are independent of its surroundings and can be considered as noise. This interference can have a noticeable effect on the depth estimation task at the pixel level.
\subsection{Analysis of CRATE layers}
\begin{figure}[thpb]
\centering
\includegraphics[scale=0.5]{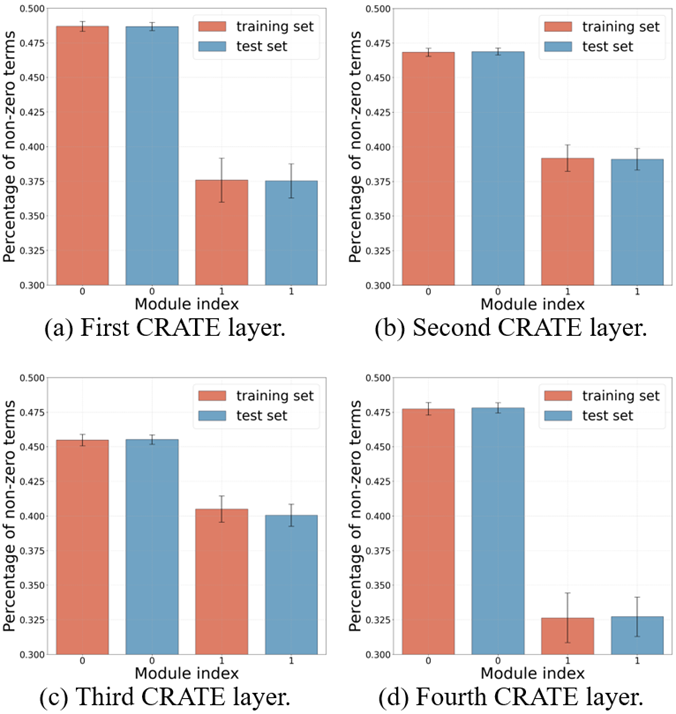}
\caption{Comparison of the percentage of non-zero elements in each CRATE module.}
\label{figurelabel5}
\end{figure}

In this subsection, experiments are conducted to verify the working mechanism of the CRATE layers. The non-zero elements and its percentage in each basic module of the CRATE layer were counted. As shown in Figure \ref{figurelabel5}, the model performed similar on the training and test sets, indicating that CRATE can efficiently implement the designed compression (denoising) and sparsification process for all data and making low-rank predictions. 
\subsection{Feature map visualization}
\begin{figure}[thpb]
\centering
\includegraphics[scale=0.5]{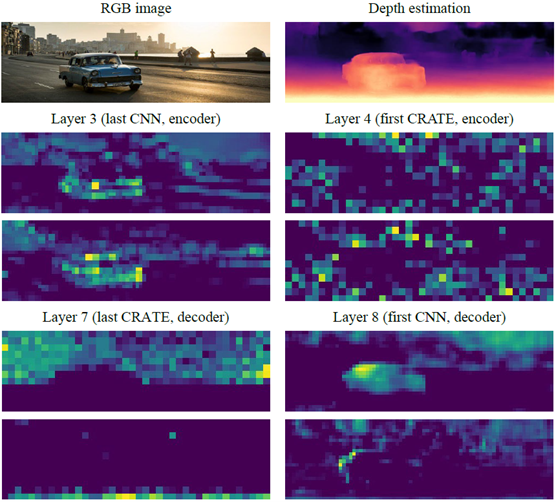}
\caption{Visualization of feature maps from CNNs and CRATE.}
\label{figurelabel6}
\end{figure}

In order to comprehensively analyze the performance and characteristics of the CCDepth model, some feature maps from 4 layers are presented to summarize the characteristics of both networks. The results are illustrated in Figure \ref{figurelabel6}. It is evident that the CNN networks (layer 3 and layer 8) emphasize more on fine details. In contrast, the CRATE network focuses more on describing the global structure. This further affirms the effectiveness of the proposed network.

\section{Conclusion}
In this paper, a novel depth estimation network, CCDepth, is proposed, which leverages CNNs and CRATE to extract the local and global information, thereby efficiently capturing useful features and improving performance in self-supervised depth estimation. Compared to current self-supervised deep estimation models, the proposed network features high estimation accuracy, significantly reduced model size, and enhanced interpretability. Nevertheless, the proposed model has relatively high time complexity due to the involvement of the CRATE layers. Future works will further improve both accuracy and efficiency.


\begin{thebibliography}{99}

\bibitem{c1} A. Masoumian, H. A. Rashwan, J. Cristiano, M. S. Asif, and D. Puig, “Monocular depth estimation using deep learning: A review,” Sensors, vol. 22, no. 14, p. 5353, 2022.
\bibitem{c2} S. Jia, S. C. Wong, and W. Wong, “Uncertainty estimation of connected vehicle penetration rate,” Transportation Science, vol. 57, no. 5, pp. 1160–1176, 2023.
\bibitem{c3} S. Jia, Y. Cai, X. Pei, Z. Yang, W. Wong, and S. C. Wong, “Exploitation of string stability to predict disturbance-triggered platoon collisions in mixed traffic comprising automated and conventional vehicles,” in 2023 IEEE 26th International Conference on Intelligent Transportation Systems (ITSC). IEEE, 2023, pp. 121–126.
\bibitem{c4} M. Alam, M. D. Samad, L. Vidyaratne, A. Glandon, and K. M. Iftekharuddin, “Survey on deep neural networks in speech and vision systems,” Neurocomputing, vol. 417, pp. 302–321, 2020.
\bibitem{c5} J. Valentin, A. Kowdle, J. T. Barron, N. Wadhwa, M. Dzitsiuk, M. Schoenberg, V. Verma, A. Csaszar, E. Turner, I. Dryanovski et al., “Depth from motion for smartphone ar,” ACM Transactions on Graphics (ToG), vol. 37, no. 6, pp. 1–19, 2018.
\bibitem{c6} W. Huang, J. Cheng, Y. Yang, and G. Guo, “An improved deep convolutional neural network with multi-scale information for bearing fault diagnosis,” Neurocomputing, vol. 359, pp. 77–92, 2019.
\bibitem{c7} X. Yang, H. Luo, Y. Wu, Y. Gao, C. Liao, and K.-T. Cheng, “Reactive obstacle avoidance of monocular quadrotors with online adapted depth prediction network,” Neurocomputing, vol. 325, pp. 142–158, 2019.
\bibitem{c8} L. Engstrom, B. Tran, D. Tsipras, L. Schmidt, and A. Madry, “Exploring the landscape of spatial robustness,” in International conference on machine learning. PMLR, 2019, pp. 1802–1811.
\bibitem{c9} C. Szegedy, W. Zaremba, I. Sutskever, J. Bruna, D. Erhan, I. Goodfellow, and R. Fergus, “Intriguing properties of neural networks,” arXiv preprint arXiv:1312.6199, 2013.
\bibitem{c10} A. Saxena, S. H. Chung, and A. Y. Ng, “3-d depth reconstruction from a single still image,” International journal of computer vision, vol. 76, pp. 53–69, 2008.
\bibitem{c11} S. Choi, D. Min, B. Ham, Y. Kim, C. Oh, and K. Sohn, “Depth analogy: Data-driven approach for single image depth estimation using gradient samples,” IEEE Transactions on Image Processing, vol. 24, no. 12, pp. 5953–5966, 2015.
\bibitem{c12} R. Furukawa, R. Sagawa, and H. Kawasaki, “Depth estimation using structured light flow–analysis of projected pattern flow on an object’s surface,” in Proceedings of the IEEE international conference on computer vision, 2017, pp. 4640–4648.
\bibitem{c13} P. Weng, S. Jia, X. Pei, and Y. Yue, “Bayes neural network with a novel pictorial feature for transportation mode recognition based on gps trajectories,” in CICTP 2021, 2021, pp. 1635–1645.
\bibitem{c14} S. Jia, Y. Yue, Z. Yang, X. Pei, and Y. Wang, “Travelling modes recognition via bayes neural network with bayes by backprop algorithm,” in CICTP 2020, 2020, pp. 3994–4004.
\bibitem{c15} R. Li, Z. Yang, X. Pei, Y. Yue, S. Jia, C. Han, and Z. He, “A novel one-stage approach for pointwise transportation mode identification inspired by point cloud processing,” Transportation Research Part C: Emerging Technologies, vol. 152, p. 104127, 2023.
\bibitem{c16} S. Jia, Y. Yue, Z. Yang, X. Pei, and Y. Lu, “A novel approach to reveal travel patterns of city residents based on mobile internet locating data,” in CICTP 2019, 2019, pp. 6146–6156.
\bibitem{c17} D. Eigen, C. Puhrsch, and R. Fergus, “Depth map prediction from a single image using a multi-scale deep network,” Advances in neural information processing systems, vol. 27, 2014.
\bibitem{c18} K. He, X. Zhang, S. Ren, and J. Sun, “Deep residual learning for image recognition,” in Proceedings of the IEEE conference on computer vision and pattern recognition, 2016, pp. 770–778.
\bibitem{c19} J. You, S. Jia, X. Pei, and D. Yao, “Dmrvisnet: Deep multihead regression network for pixel-wise visibility estimation under foggy weather,” IEEE Transactions on Intelligent Transportation Systems, vol. 23, no. 11, pp. 22 354–22 366, 2022.
\bibitem{c20} D. Wang, S. Jia, X. Pei, C. Han, D. Yao, and D. Liu, “Dernet: Driver emotion recognition using onboard camera,” IEEE Intelligent Transportation Systems Magazine, 2023.
\bibitem{c21} C. Godard, O. Mac Aodha, and G. J. Brostow, “Unsupervised monocular depth estimation with left-right consistency,” in Proceedings of the IEEE conference on computer vision and pattern recognition, 2017, pp. 270–279.
\bibitem{c22} C. Godard, O. Mac Aodha, M. Firman, and G. J. Brostow, “Digging into self-supervised monocular depth estimation,” in Proceedings of the IEEE/CVF international conference on computer vision, 2019, pp. 3828–3838.
\bibitem{c23} S. Jia, X. Pei, W. Yao, and S. C. Wong, “Self-supervised depth estimation leveraging global perception and geometric smoothness,” IEEE Transactions on Intelligent Transportation Systems, vol. 24, no. 2, pp. 1502–1517, 2022.
\bibitem{c24} S. Jia, X. Pei, Z. Yang, S. Tian, and Y. Yue, “Novel hybrid neural network for dense depth estimation using on-board monocular images,” Transportation research record, vol. 2674, no. 12, pp. 312–323, 2020.
\bibitem{c25} S. Jia, X. Pei, X. Jing, and D. Yao, “Self-supervised 3d reconstruction and ego-motion estimation via on-board monocular video,” IEEE Transactions on Intelligent Transportation Systems, vol. 23, no. 7, pp. 7557–7569, 2021.
\bibitem{c26} S. Jia and W. Yao, “Self-supervised multi-task learning framework for safety and health-oriented road environment surveillance based on connected vehicle visual perception,” International Journal of Applied Earth Observation and Geoinformation, vol. 128, p. 103753, 2024.
\bibitem{c27} M. Poggi, F. Aleotti, F. Tosi, and S. Mattoccia, “Towards real-time unsupervised monocular depth estimation on cpu,” in 2018 IEEE/RSJ international conference on intelligent robots and systems (IROS). IEEE, 2018, pp. 5848–5854.
\bibitem{c28} J. Liu, Q. Li, R. Cao, W. Tang, and G. Qiu, “Mininet: An extremely lightweight convolutional neural network for real-time unsupervised monocular depth estimation,” ISPRS Journal of Photogrammetry and Remote Sensing, vol. 166, pp. 255–267, 2020.
\bibitem{c29} L. Huynh, P. Nguyen-Ha, J. Matas, E. Rahtu, and J. Heikkila, “Guiding ¨ monocular depth estimation using depth-attention volume,” in Computer Vision–ECCV 2020: 16th European Conference, Glasgow, UK, August 23–28, 2020, Proceedings, Part XXVI 16. Springer, 2020, pp. 581–597.
\bibitem{c30} S. Jia and W. Yao, “Joint learning of frequency and spatial domains for dense image prediction,” ISPRS Journal of Photogrammetry and Remote Sensing, vol. 195, pp. 14–28, 2023.
\bibitem{c31} Y. Yu, S. Buchanan, D. Pai, T. Chu, Z. Wu, S. Tong, B. Haeffele, and Y. Ma, “White-box transformers via sparse rate reduction,” Advances in Neural Information Processing Systems, vol. 36, 2024.
\bibitem{c32} T. Zhou, M. Brown, N. Snavely, and D. G. Lowe, “Unsupervised learning of depth and ego-motion from video,” in Proceedings of the IEEE conference on computer vision and pattern recognition, 2017, pp. 1851–1858.
\bibitem{c33} Z. Yin and J. Shi, “Geonet: Unsupervised learning of dense depth, optical flow and camera pose,” in Proceedings of the IEEE conference on computer vision and pattern recognition, 2018, pp. 1983–1992.
\bibitem{c34} V. Casser, S. Pirk, R. Mahjourian, and A. Angelova, “Depth prediction without the sensors: Leveraging structure for unsupervised learning from monocular videos,” in Proceedings of the AAAI conference on artificial intelligence, vol. 33, no. 01, 2019, pp. 8001–8008.
\bibitem{c35} A. Geiger, P. Lenz, C. Stiller, and R. Urtasun, “Vision meets robotics: The kitti dataset,” The International Journal of Robotics Research, vol. 32, no. 11, pp. 1231–1237, 2013.
\bibitem{c36} D. Eigen and R. Fergus, “Predicting depth, surface normals and semantic labels with a common multi-scale convolutional architecture,” in Proceedings of the IEEE international conference on computer vision, 2015, pp. 2650–2658.
\bibitem{c37} D. P. Kingma and J. Ba, “Adam: A method for stochastic optimization,” arXiv preprint arXiv:1412.6980, 2014.

\end{thebibliography}
\end{document}